\documentclass[letterpaper]{article} 
\usepackage{aaai25}  
\usepackage{times}  
\usepackage{helvet}  
\usepackage{courier}  
\usepackage[hyphens]{url}  
\usepackage{graphicx} 
\usepackage{natbib}  
\usepackage{caption} 
\usepackage{enumitem}  
\usepackage{newfloat}
\usepackage{listings}
\usepackage{xcolor}         
\usepackage{microtype}
\usepackage{url}
\usepackage{multirow}
\usepackage{makecell}
\usepackage{booktabs}
\usepackage{adjustbox}
\usepackage{enumitem}
\usepackage{arydshln}
\usepackage{array}
\usepackage{ragged2e}
\usepackage{xcolor}
\usepackage{booktabs}
\usepackage{comment}
\usepackage{listings}
\usepackage{inconsolata}
\usepackage{pifont}
\definecolor{darkblue}{rgb}{0, 0, 0.5}
\usepackage{afterpage}
\usepackage{graphicx}
\usepackage{algorithm}
\usepackage{algpseudocode}
\usepackage{amsmath}
\usepackage{amsthm}
\usepackage{zref-xr,zref-user}
\usepackage{amssymb}
\usepackage{mathtools}

\theoremstyle{plain}
\newtheorem{theorem}{Theorem}

\theoremstyle{definition}
\newtheorem{definition}[theorem]{Definition}

\theoremstyle{remark}
\newtheorem{remark}[theorem]{Remark}
\DeclareCaptionStyle{ruled}{labelfont=normalfont,labelsep=colon,strut=off} 
\floatstyle{ruled}
\newfloat{listing}{tb}{lst}{}
\floatname{listing}{Listing}
\title{CodeHalu: Investigating Code Hallucinations in LLMs via \\Execution-based Verification}
\author{
    Yuchen Tian\textsuperscript{\rm 1}\equalcontrib,
    ~Weixiang Yan\textsuperscript{\rm 2}\equalcontrib\footnotemark[2], 
    ~Qian Yang\textsuperscript{\rm 3 4}, 
   ~Xuandong Zhao\textsuperscript{\rm 5} \\
    Qian Chen\textsuperscript{\rm 6}, 
    ~Wen Wang\textsuperscript{\rm 6}, 
    ~Ziyang Luo\textsuperscript{\rm 1}, 
    ~Lei Ma\textsuperscript{\rm 7 8}\thanks{The corresponding author.}, 
    ~Dawn Song\textsuperscript{\rm 5}\footnotemark[2]\\
}
\affiliations{
    \textsuperscript{\rm 1}Hong Kong Baptist University~~
    \textsuperscript{\rm 2}UC Santa Barbara~~
    \textsuperscript{\rm 3}Mila - Québec AI Institute~~
    \textsuperscript{\rm 4}Université de Montréal\\
    \textsuperscript{\rm 5}UC Berkeley~~
    \textsuperscript{\rm 6}Alibaba Group~~
    \textsuperscript{\rm 7}The University of Tokyo~~
    \textsuperscript{\rm 8}University of Alberta\\
    
    yuchent@connect.hku.hk, 
    weixiangyan@ucsb.edu\\

}

\begin{document}

\maketitle

\begin{abstract}
Large Language Models (LLMs) have made significant progress in code generation, offering developers groundbreaking automated programming support. However, LLMs often generate code that is syntactically correct and even semantically plausible, but may not execute as expected or fulfill specified requirements. This phenomenon of hallucinations in the code domain has not been systematically explored. To advance the community's understanding and research on this issue, we introduce the concept of \textbf{code hallucinations} and propose a classification method for code hallucination based on execution verification.
We categorize code hallucinations into four main types: \textbf{mapping}, \textbf{naming}, \textbf{resource}, and \textbf{logic} hallucinations, with each category further divided into different subcategories to understand and address the unique challenges faced by LLMs in code generation with finer granularity.
Additionally, we present a dynamic detection algorithm called \textbf{CodeHalu} designed to detect and quantify code hallucinations. We also introduce the \textbf{CodeHaluEval} benchmark, which includes 8,883 samples from 699 tasks, to systematically and quantitatively evaluate code hallucinations. By evaluating 17 popular LLMs using this benchmark, we reveal significant differences in their accuracy and reliability in code generation, offering detailed insights for further improving the code generation capabilities of LLMs. The CodeHalu benchmark and code are publicly available at \url{https://github.com/yuchen814/CodeHalu}.
\end{abstract}

%

\section{Introduction}
\label{sec:intro}

Deep neural networks often generate erroneous information that contradicts the original content, cannot be verified, or conflicts with real-world knowledge. This phenomenon, commonly known as model hallucination, attracts widespread attention in the fields of natural language processing and multimodal learning~\citep{ji2023survey,zhang2023siren,liu2024survey}, with the community actively exploring methods to mitigate hallucinations~\citep{peng2023check,elaraby2023halo,liu2023aligning}. However, the issue of model hallucination in the code generation domain remains unexplored.

Conducting a thorough and dedicated study on code hallucinations is crucial for improving the quality of code generated by LLMs. Firstly, the purpose of code is to solve problems, and its value is realized only when the code executes successfully and passes tests~\citep{chen2021evaluating,austin2021program,yan2023codescope}. This necessitates that the generated code not only maintain strict logic and precision but also undergoes execution verification to confirm its correctness. Therefore, the practical use and verification of code differ significantly from Natural Language(NL) texts, meaning we cannot directly apply the definitions and methods used for NL hallucinations to code. Secondly, code snippets containing hallucinations may trigger runtime errors, 
or exhibit functional defects, which hinder the reliable deployment of LLMs in automated software development scenarios. Lastly, by exploring and verifying code hallucinations in a targeted manner, we can effectively uncover their causes and contribute to improving the architecture and training methods of LLMs.

To fill this gap, we define the concept of \textit{code hallucination} in LLMs, based on the unique purpose and function of the code. \textbf{Code hallucinations refer to the phenomenon where code generated by LLMs is syntactically correct or even semantically plausible but ultimately cannot execute as expected or fails to meet specified requirements.}\footnote{We test 16 LLMs using 105,958 code samples. The experimental results demonstrate that only 9 models occasionally exhibit syntactic errors in the generated code, with an exceptionally low average error rate of \textbf{0.0020}. These findings support our initial hypothesis that the code generated by LLMs is generally syntactically correct and even semantically plausible or appropriate. 
} This phenomenon typically arises from various factors, such as errors or outdated information in the training data, an inadequate grasp of the syntax rules and programming paradigms of the programming languages, and limitations in the logical processing capabilities of the models.
In contrast to previous methods that \textit{passively} explore hallucinations in NLP through a Q\&A framework or by prompting LLMs to generate hallucinated answers~\citep{lin2021truthfulqa,cheng2023evaluating}, we employ an \textit{active} strategy to detect hallucinations during the code generation process by LLMs. This approach is crucial as the ultimate goal of the generated code is to execute correctly and fulfill specific tasks.

\begin{figure*}[t]
    \centering
    \hspace{-19pt}
    \includegraphics[width=\textwidth]{figures/ch.pdf}
    \caption{The definition and classification of code hallucinations, including 4 main categories and 8 subcategories.}
    \label{fig:summary}
\end{figure*}

To detect and quantify hallucinations in LLMs during code generation, we develop a dynamic detection algorithm named \textbf{CodeHalu}. This algorithm employs a statistical induction method based on execution validation to identify specific patterns that frequently occur in code generated by multiple LLMs, such as \textit{error types}, \textit{syntax interruptions}, or \textit{unexpected execution results}. When a pattern consistently appears across multiple LLMs, it is recognized as a common code hallucination. Based on the CodeHalu algorithm \ref{algo:codehalu_algo}, we employ an execution-based validation approach for hallucination detection, combined with a two-stage heuristic identification method. By conducting statistical quantification on 17 mainstream LLMs, we categorize code hallucinations into four major categories: \textbf{Mapping}, \textbf{Naming}, \textbf{Resource}, and \textbf{Logical Hallucinations}. These categories are further divided into eight subcategories, as illustrated in Figure~\ref{fig:summary}. We analyze 17 LLMs for cross-task occurrence rates in eight categories of code hallucinations. The low average rate of 2.04\% confirms the independence and validity of our classification.

To effectively measure and compare code hallucinations across different LLMs, we introduce an evaluation benchmark named \textbf{CodeHaluEval}, which is based on the incidence rate of hallucinations. It follows a structured process of \textit{Validation-Identification-Construction} as shown in Figure~\ref{fig:codehalu_process} to detect and evaluate code hallucinations in LLMs, closely tied to real-world programming scenarios, ensuring that the generated code correctly achieves the expected functionality. CodeHaluEval encompasses eight types of code hallucinations as illustrated in Figure~\ref{fig:summary}, covering 699 distinct tasks and corresponding to 8,883 samples. Additionally, we systematically evaluate 17 mainstream LLMs to reveal the distribution and behavior patterns of their code hallucinations. We also analyze the potential causes of various code hallucinations, providing detailed insights for further improving the code generation capabilities of LLMs. Our contributions can be summarized as follows:

\begin{itemize}[leftmargin=*, itemsep=0pt, topsep=0pt]

\item \textbf{Code Hallucination}: We introduce the concept of \textit{code hallucination} in LLMs and propose an execution-based verification method to define code hallucination, addressing a gap in the research on hallucination within the code generation domain.

\item \textbf{CodeHalu Algorithm}: We develop a dynamic detection algorithm, CodeHalu, to identify and quantify the types of hallucinations that occur in LLMs during code generation. We categorize code hallucinations into four main categories based on a two-stage heuristic approach, discussing their theoretical implications and potential causes.

\item \textbf{CodeHaluEval Benchmark}: We propose the CodeHaluEval benchmark to systematically evaluate 17 popular LLMs, revealing the distribution and patterns of code hallucinations across these models, and providing insights for developing more robust and reliable LLMs.

\end{itemize}

\section{Related Work}
\label{sec:related_work}

\subsection{Hallucination}
\label{sec:lm_mm_hallucination}

In the field of NLP, hallucination is initially defined as the phenomenon where the text generated by a model is fluent and natural but either lacks substantive meaning or is inconsistent with the provided source content~\citep{ji2023survey}. Recently, \citet{zhang2023siren} standardize the definition of NL hallucinations in LLMs into three categories: \textit{input-conflicting hallucinations}, where the content generated by LLMs diverges from the user's input; \textit{context-conflicting hallucinations}, in which the generated content contradicts previously generated content; and \textit{fact-conflicting hallucinations}, where the generated content conflicts with established world knowledge.
These hallucinations are attributed to various factors, such as poor-quality data samples in the training dataset or the use of sampling algorithms with high uncertainty.

In the multimodal domain, \citet{zhai2023halle} classify types of hallucinations in image-to-text scenarios, such as image captioning and visual question answering. They define three main types of hallucinations: \textit{object existence hallucinations}, \textit{object attribute hallucinations}, and \textit{object relationship hallucinations}. In text-to-image scenarios, such as image generation, hallucinations refer to the creation of factually incorrect details by the image generation model in response to the given text input. \citet{huang2024visual} introduce VHTest, which evaluates hallucinations across eight dimensions in images, including the \textit{existence}, \textit{shape}, \textit{color}, \textit{orientation}, \textit{OCR}, \textit{size}, \textit{position}, and \textit{counting} of visual objects. 
In text-to-video scenarios, such as video generation, \citet{chu2024sora} define three types of hallucinations: \textit{prompt consistency hallucinations}, \textit{static hallucinations}, and \textit{dynamic hallucinations}. 
\emph{Although the issue of hallucinations receives extensive attention in NLP and multimodal domains, it remains unexplored in the code domain. Therefore, we propose CodeHalu to systematically define, identify, classify, and quantify code hallucinations in LLMs.}

\subsection{Existing Coding Benchmarks}
\label{sec:code_bench}
In recent years, numerous studies focus on evaluating the capability of LLMs to handle various programming tasks. Among these, the HumanEval~\citep{chen2021evaluating}, includes 164 Python programming problems, each with an average of 6.7 unit tests. 
The APPS~\citep{hendrycks2021measuring} benchmark presents more challenging programming questions, with each problem averaging 293.2 words in length. CodeScope~\citep{yan2023codescope} covers 43 programming languages and 8 coding tasks to evaluate LLMs in code understanding and generation. 
MMCode~\citep{li2024mmcode} is designed to evaluate the programming capability of code models in multimodal scenarios.
SWE-bench~\citep{jimenez2023swe} evaluates the ability of LLMs to modify code repositories and solve problems with a complexity level comparable to what human programmers encounter.
\emph{Overall, existing code benchmarks focus on evaluating the performance of LLMs on various programming tasks. However, there is still a lack of effective methods to detect and quantify potential hallucinations that may occur in code generation. Therefore, we propose CodeHaluEval to detect and quantify code hallucinations in LLMs.}

\section{Code Hallucination}
\label{sec:code_hallucination}

As a tool, code aims to achieve specific objectives through correct execution. This inherent characteristic motivates our use of an execution-based verification method to explore and identify code hallucinations. In this section, we define the concept of \textit{code hallucination} and distinguish it from code errors, clarifying the relationship and differences between these two phenomena.

\begin{definition}[Code Hallucinations]
Code hallucinations refer to the code generated by large language models that is syntactically correct or even semantically plausible, but ultimately cannot execute as expected or fails to meet specified requirements.
\end{definition}

\begin{definition}[Code Errors]
Code errors refer to issues in a program that cause it to stop executing.
\end{definition}

\begin{figure}[t]
\centering
\includegraphics[width=0.48\textwidth]{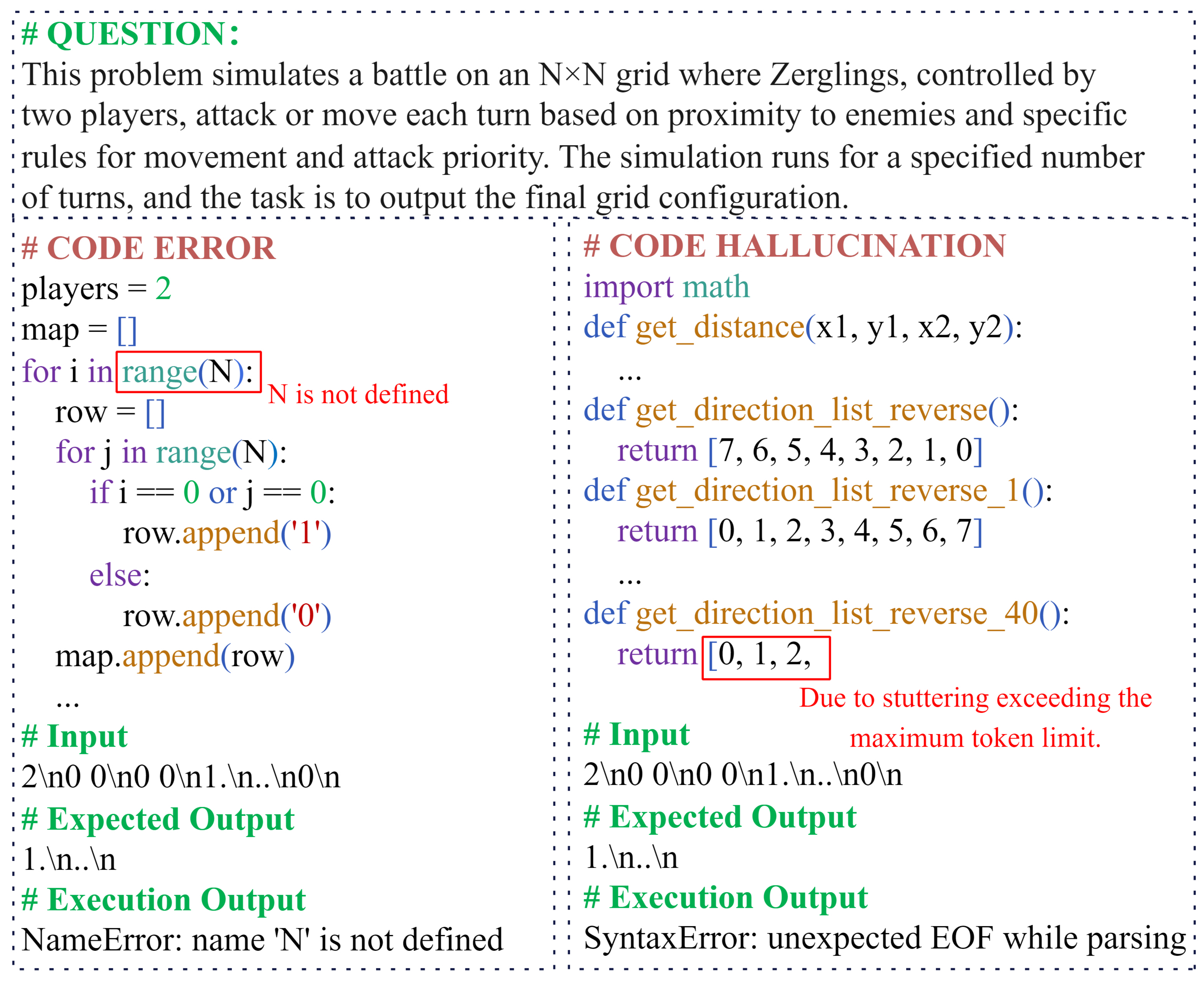}
\caption{
Examples that differentiate between code errors and code hallucinations.
}
\label{fig:data_example}
\end{figure}

\begin{remark}[Code Hallucinations vs. Code Errors]

In multiple domains, existing work~\citep{ji2023survey,zhang2023siren,huang2024visual,zhai2023halle,chu2024sora} often equates errors with hallucinations, or considers errors as a specific subset of hallucinations. We follow this perspective and regard code errors as a specific subset of code hallucinations. In other words, errors manifest as a form of hallucination, but not all hallucinations can be adequately expressed through errors. Figure \ref{fig:data_example}  illustrates the distinction between code errors and code hallucinations. The code on the left exhibits a typical code error due to the use of an undefined variable ``N'', resulting in a NameError. On the right, the code repeatedly calls the same function due to a logical collapse during generation, eventually exceeding the maximum token limit and leading to a SyntaxError. However, the underlying issue is a latent logical hallucination, rather than the observed syntactic error.
\end{remark}

Overall, although there is some slightly overlap between code hallucinations and code errors, their meanings, research objects, and scopes differ significantly. Code hallucinations focus on why the model produces hallucinations, while code errors focus on what grammatical rules the code violates. Code errors form a proper subset of code hallucinations, while code hallucinations encompass a broader range of potential logical and functional issues, representing a finer-grained and more comprehensive evaluation of the overall quality and functionality of the code.

\section{CodeHalu Algorithm}
\label{sec:codehalu_algo}

\begin{algorithm}[t]
\caption{CodeHalu Algorithm}
\label{algo:codehalu_algo}
\begin{algorithmic}[1]
\small
\Statex \hspace{-18pt} \textbf{Input:} Code Generation Dataset $\alpha$, Language models $\mathsf{\pi}$
\Statex \hspace{-18pt} \textbf{Output:} HaluTypes $\xi$
\State Let $\xi \gets$ empty list

\State \textbf{for} $\alpha_i$, where $i \in \{1, \dots, k\}$ \textbf{do}
\State \hspace{1em} \textbf{for} $\pi_j$, where $j \in \{1, \dots, m\}$ \textbf{do}
\State \hspace{2em} $\mathsf{GC}_j^{\alpha_i} \gets$ $\pi_j(\mathsf{GI}_j, \mathsf{Q})$
\State \hspace{2em} \textbf{if} $\mathsf{GC}_j^{\alpha_i}$ is stuttering, infinite enumeration, or gibberish
\State \hspace{3em} $\xi\leftarrow \xi \cup \operatorname{State}(\mathsf{GC}_j^{\alpha_i})$
\State \hspace{2em} \textbf{else}
\State \hspace{3em} \textbf{for} $t_n$, where $n \in \{1, \dots, N\}$ \textbf{do}
\State \hspace{4em} \textbf{if} \text{Execute}($\mathsf{GC}_j^{\alpha_i}(t_n))$
\State \hspace{5em} $\mathsf{ER}_j^{\alpha_i}(t_n) \gets $\text{Execute}$(\mathsf{GC}_j^{\alpha_i}(t_n))$
\State \hspace{5em} \textbf{if} $\mathsf{ER}_j^{\alpha_i}(t_n) \neq op_{t_n}$
\State \hspace{6em} $\xi\leftarrow \xi \cup \operatorname{State}(\mathsf{GC}_j^{\alpha_i})$
\State Aggregate and count frequencies of unique $\operatorname{State}(\mathsf{GC}_j^{\alpha_i})$ in $\xi$

\end{algorithmic}
\end{algorithm}



In this section, we introduce a dynamic detection algorithm called \textbf{CodeHalu}, which detects and quantifies hallucinations in LLMs in code generation. CodeHalu operates on the assumption \(\mathsf{ASS}\): if a specific pattern frequently appears in the code generated by multiple LLMs, it is considered a common code hallucination. These patterns include \textit{error types}, \textit{syntax interruptions}, \textit{logical collapse}, or \textit{unexpected execution results}.

Consider a dataset $\alpha$ contains $k$ samples, where each sample $\alpha_i$ consists of a problem description $Q$ and a series of test cases $t_1, t_2, \ldots, t_n$. Each test case $t_n$ includes an input $\mathsf{ip}$ and the corresponding expected output $\mathsf{op}$. Notably, following previous work~\citep{li2023taco,yan2023codescope}, we integrate resource (time and memory) constraints into the code generation instructions. As shown in Algorithm \ref{algo:codehalu_algo}, we use a \(\mathsf{\pi}_j\) to generate a code solution \(\mathsf{GC}_j^{\alpha_i}\) for each sample $\alpha_i$ based on the code generation instruction $\mathsf{GI_j}$ and the problem description $\mathsf{Q}$. If $\mathsf{GC}_j^{\alpha_i}$ exhibits any of the states such as stuttering, infinite loops, or gibberish, we include it in $\xi$.

To test the potential hallucinations of $\mathsf{GC}_j^{\alpha_i}$ at a fine-grained level, we execute all test cases of sample \(\alpha_i\) one by one to verify whether it successfully executes and meets the expected functionality. We record the actual execution result \(\mathsf{ER}_j^{\alpha_i}(t_n)\) of the code under each test case \(t_n\) and extensively test each sample \(\alpha_i\) across more than 15 $\mathsf{\pi}$ to obtain statistically-based inductive results. If the code execution fails or does not meet the expected results, we record it in $\xi$.

Finally, we merge the identical states \(\operatorname{State}(\mathsf{GC}_j^{\alpha_i})\) detected by CodeHalu and calculate their occurrence frequencies. According to assumption \(\mathsf{ASS}\), $\xi$ can be represented as [($\xi_1$, $P_1$), ($\xi_2$, $P_2$), \ldots, ($\xi_o$, $P_o$)], where $\xi_o$ denotes the \(o^{th}\) type of code hallucination, and \(P_o\) indicates its corresponding frequency. \textbf{Code hallucinations come from four perspectives: errors, syntax, logic, and execution results}. Additionally, CodeHalu is language-agnostic and can dynamically adjust to match various programming scenarios depending on the programming language.

\section{Code Hallucinations Classification}
\label{sec:classification_definitions}

In this section, we analyze the hallucination states detected by the CodeHalu algorithm, classify and define four main types of hallucinations, and discuss the rationale behind the classification method.

According to the TIOBE Index\footnote{\url{https://www.tiobe.com/tiobe-index/}}, a metric of programming language popularity, we primarily investigate code hallucinations within the Python. By applying the CodeHalu algorithm on the complex APPS dataset~\citep{hendrycks2021measuring} and 17 widely-used LLMs, we identify and validate 18 types of hallucination states that violate human expectations during the code generation, including \textit{inconsistent code context}, \textit{ambiguous logic and data flow}, \textit{conflicting intentions}, among others. Using the two-stage heuristic classification method introduced in Remark~\ref{sec:classification_discuss}, we categorize code hallucinations into four main types based on the nature and origin of these phenomena: mapping hallucinations, naming hallucinations, resource hallucinations, and logical hallucinations, as illustrated in Figure \ref{fig:summary}.

\begin{definition}[Mapping Hallucinations]
Mapping Hallucinations refer to the ambiguity and confusion that occur in LLMs' perception and mapping of data types, values, and structures during data operations. This phenomenon is further divided into two sub-categories: \textit{data compliance hallucinations} and \textit{structure access hallucinations}.

\textit{Data compliance hallucinations} occur when LLMs have a vague understanding of the data types and parameter values of the objects being manipulated, resulting in generated code that attempts to perform type-mismatched or rule-violating operations.

\textit{Structure access hallucinations} occur when LLMs misinterpret the data structures of the objects being manipulated, leading to generated code that attempts to access non-existent array indices or dictionary keys.

\end{definition}

\begin{definition}[Naming Hallucinations]
    
Naming Hallucinations refer to the memory-related issues and factual inaccuracies exhibited by LLMs when handling the naming, scope, and existence of variables, attributes, and modules. This phenomenon is further divided into two subcategories: \textit{identity hallucinations} and \textit{external source hallucinations}.

\textit{Identity hallucinations} occur when LLMs possess biased memories or lack sufficient understanding of the context, leading to generated code that references undefined variables, accesses non-existent object properties, or uses unassigned variables in local scopes. 

\textit{External source hallucinations} occur when LLMs exhibit significant memory-related issues or obvious conflicts with facts concerning external knowledge sources, resulting in generated code that attempts to import non-existent modules or fails to correctly load modules from other paths. 
\end{definition} 

\begin{definition} [Resource Hallucinations] 
Resource Hallucinations occur when LLMs lack an adequate perception and prediction of resource consumption and control flow of the generated code during execution. This phenomenon is further divided into \textit{physical constraint hallucinations} and \textit{computational boundary hallucinations}.

\textit{Physical constraint hallucinations} arise when LLMs underestimate resource consumption during data processing operations, causing code failure due to exceeding memory capacity, stack depth, or other physical constraints. 

\textit{Computational boundary hallucinations} occur when LLMs blur recognition of numerical calculation limits and iteration endpoints during data processing operations, causing code failure due to numerical overflow or improper iteration control. 
\end{definition}

\begin{definition}[Logic Hallucinations] 
Logic Hallucinations refer to the discrepancies between the expected results and the actual outcomes after executing the code generated by LLMs, or outputs with low semantic density or even complete chaos. This phenomenon is further divided into \textit{logic deviation} and \textit{logic breakdown}.

\textit{Logic deviation} occurs when LLMs generate code that lacks sufficient logical consideration or contradicts the intended instructions. While this hallucination may not cause errors during execution, logical deviations or confusion result in outcomes that fail to meet the expected results.

\textit{Logic breakdown} occurs when LLMs struggle to interpret or maintain a continuous understanding of context during code generation. This indicates that the models may lose direction while generating code, making it difficult to maintain strict consistency of contextual information. 
\end{definition}

\begin{figure}[t]
\centering
\includegraphics[width=0.48\textwidth]{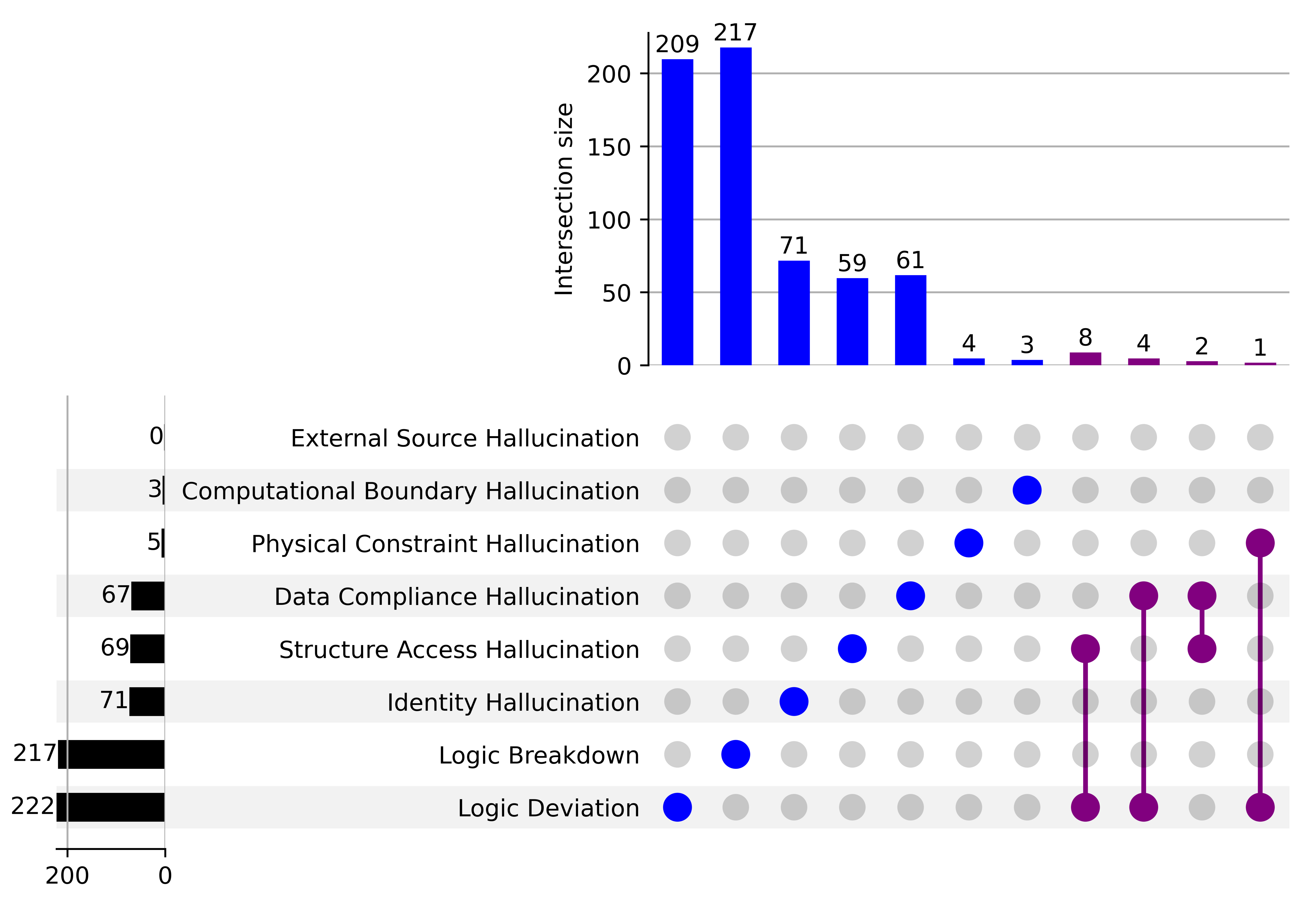}
\caption{The diagram illustrates the intersection of various hallucinations in Gemma-7B during the CodeHaluEval. The bar chart at the top shows the frequency of each intersection, while the bar chart on the left indicates the frequency of each type of hallucination. The connecting lines represent the co-occurrence patterns between different hallucinations.}
\label{fig:gemma_corss_rate}
\end{figure}

\begin{remark}[Discussion of Rationality]
\label{sec:classification_discuss}

To ensure the rationality and effectiveness of our code hallucination classification method, we conduct in-depth analyses.

Firstly, we extensively reference classification methods for hallucinations in the fields of NLP and multimodal research~\citep{zhang2023siren,ji2023survey,huang2024visual,zhai2023halle,chu2024sora}, as well as methods for classifying code errors and vulnerabilities in software engineering~\citep{pan2023understanding,wang2024large,10298532}. We adopt a two-stage heuristic classification strategy. Initially, team members independently review failed code cases and develop preliminary classification frameworks; then, we reach a consensus through collaborative discussions. This widely used approach ensures the adaptability and accuracy of our framework, enabling a systematic understanding of code hallucinations in LLMs.

Secondly, we analyze the cross-task occurrence rates of each model across eight categories of code hallucinations.
The results show that the average cross-task occurrence rate for these categories is only 2.04\%, confirming the independence and rationality of our classification. For the Gemma-7B model, which exhibits the most severe hallucinations in Table \ref{table:bench}, only 1.07\% of task samples show cross-task hallucinations, as illustrated in Figure \ref{fig:gemma_corss_rate}.

Lastly, we conduct an empirical investigation of our classification results and design a questionnaire to evaluate the rationality of our method
. The survey receives 23 responses, and after excluding seven respondents with less than three years of experience, we analyzed 16 valid responses. The survey results indicate a rationality rating of 91.08\% for our classification method, further supporting its validity.

\end{remark}

\section{Cause Analysis of Code Hallucinations}
\label{sec:cause_analysis}
In this section, we explore the potential causes of various hallucinations generated by LLMs, aiming to provide valuable insights for optimizing training data, training methods, model architecture, and alignment strategies.

\textbf{Mapping hallucinations} primarily stem from the model's misunderstanding of data types and structures. This phenomenon arises due to several factors: (1) The model generates code based on tokens, lacking insight into higher-level structures such as statements and functions~\citep{e23091174}; (2) When dealing with long-distance data dependencies, especially within complex code blocks, the model fails to continuously track the structure and state of variables, overly relying on local information while neglecting the importance of the overall context~\citep{zhang2024hirope}; (3) The model does not explicitly perform type checking and structure matching during code generation, lacking static checking and error correction mechanisms.

\textbf{Naming hallucinations} reflect the limitations of models in tracking information and utilizing external knowledge. This issue arises from several factors: (1) Token-based feature representation makes it difficult to accurately model long-distance dependencies, leading to model misjudgments regarding variable scope, lifecycle, and visibility~\citep{xu-etal-2020-learning}; (2) The code generation process lacks consistency checks for identifiers and does not perform global tracking of variable definitions and usage; (3) Knowledge of external libraries is not effectively and timely integrated into the model's knowledge system, making it difficult for the model to accurately understand the names, functions, and invocation methods of libraries~\citep{10174227}.

\textbf{Resource hallucinations} highlight the model's lack of deep understanding of code execution mechanisms and physical constraints. These issues arise from several factors: (1) The training data lacks information related to resource consumption and performance optimization, making it difficult for the model to learn about complexity analysis and resource estimation; (2) As the model generates code based on probabilities, it lacks a module for calculating and estimating the resource consumption of the generated code, making it unable to simulate the real-world operating environment and resource limits; (3) During the model training process, the focus is usually on the correctness of the code's functionality, often overlooking its complexity and resource constraints in actual execution environments.

\begin{figure*}[t]
    \centering
    \includegraphics[width=\textwidth]{figures/codehalu_processing.pdf}
    \caption{Collection of CodeHaluEval benchmark based on a \textbf{verification-identification-construction} process.}
    \label{fig:codehalu_process}
\end{figure*}

\textbf{Logic hallucinations} reveal the model's deficiencies in semantic understanding and reasoning about code. This issue arises due to several factors: (1) The model mainly relies on pattern matching and statistical rules to generate code, lacking a fundamental understanding of symbolic systems and rigorous verification of program logic; (2) The training data is often not rigorously verified for accuracy and may contain code with very similar functions. Since models sometimes imitate and memorize previous examples~\citep{yan2022whygen}, this can result in the model directly replicating similar logic in the code or even learning incorrect logic from the outset; (3) When the model generates code, repetition at the line level has a self-reinforcing effect, causing the model to become increasingly confident in the code it generates, which may lead to a stuttering phenomenon~\citep{xu2022learning}.

\section{The CodeHaluEval Benchmark}
\label{sec:benchmark}

We construct the \textbf{CodeHaluEval} benchmark, a unified evaluation method for comparing various types and frequencies of hallucinations in code generation across different LLMs. We develop the CodeHaluEval benchmark based on the APPS testing set, following a structured process of \textbf{Validation-Identification-Construction}, as shown in Figure \ref{fig:codehalu_process}.

\begin{table}[t]
\centering
    \begin{adjustbox}{width=0.47\textwidth}
     \renewcommand{\arraystretch}{1.2}
    \begin{tabular}{cccccc}
    \toprule
     \textbf{Category} & \textbf{\#Tasks}  & \textbf{\#Samples} &   \textbf{Sub-Category} & \textbf{\#Tasks}  & \textbf{\#Samples} \\
    \midrule
    \multirow{2}{*}{\textbf{Mapping}} & \multirow{2}{*}{262} & \multirow{2}{*}{2,288} & \multirow{0.9}{*}{\textbf{Data Compliance}} & \multirow{0.9}{*}{110} & \multirow{0.9}{*}{941} \\
    \cline{4-6}
    &  & & \multirow{1.4}{*}{\textbf{Structure Access}} & \multirow{1.4}{*}{152} & \multirow{1.4}{*}{1347} \\
\midrule
    \multirow{2}{*}{\textbf{Naming}} & \multirow{2}{*}{157} & \multirow{2}{*}{1,853} & \multirow{0.9}{*}{\textbf{Identity}} & \multirow{0.9}{*}{115} & \multirow{0.9}{*}{1323} \\
        \cline{4-6}
    &  & & \multirow{1.4}{*}{\textbf{External Source}} & \multirow{1.4}{*}{42} & \multirow{1.4}{*}{530}\\
\midrule
    \multirow{2}{*}{\textbf{Resource}} & \multirow{2}{*}{107} & \multirow{2}{*}{1,130} & \multirow{0.9}{*}{\textbf{Physical Constraint}} & \multirow{0.9}{*}{47} & \multirow{0.9}{*}{491} \\
        \cline{4-6}
    &  & & \multirow{1.4}{*}{\textbf{Computational Boundary}} & \multirow{1.4}{*}{60} & \multirow{1.4}{*}{639} \\
\midrule
    \multirow{2}{*}{\textbf{Logic}} & \multirow{2}{*}{173} & \multirow{2}{*}{3,612} & \multirow{0.9}{*}{\textbf{Logic Deviation}} & \multirow{0.9}{*}{119} & \multirow{0.9}{*}{2,443} \\
        \cline{4-6}
 &  & & \multirow{1.4}{*}{\textbf{Logic Breakdown}} & \multirow{1.4}{*}{54} & \multirow{1.4}{*}{1,169}\\
    \bottomrule
  \end{tabular}
  \end{adjustbox}
\caption{Detailed statistics of categories, and quantities in \textbf{CodeHaluEval} benchmark.}
      \label{table:codehalueval}
\end{table}

\begin{table*}[ht]
\centering
\begin{adjustbox}{width=\textwidth}
\begin{tabular}{lccccccccccccccccc}
\toprule
   \multirow{2.3}{*}{\textbf{Model}} & \multicolumn{3}{c}{\textbf{Mapping ($\downarrow$)}} & &   \multicolumn{3}{c}{\textbf{Naming ($\downarrow$)}} &  &  \multicolumn{3}{c}{\textbf{Resource ($\downarrow$)}} &  &  \multicolumn{3}{c}{\textbf{Logic ($\downarrow$)}}& &  \multirow{2}{*}{\textbf{Average ($\downarrow$)}} \\
   \cline{2-4} \cline{6-8} \cline{10-12} \cline{14-16}
    & \multirow{1.4}{*}{\textbf{DC}}  &  \multirow{1.4}{*}{\textbf{SA}} & \multirow{1.4}{*}{\textbf{Avg.}} & & \multirow{1.4}{*}{\textbf{ID}}  & \multirow{1.4}{*}{\textbf{ES}} & \multirow{1.4}{*}{\textbf{Avg.}} & & \multirow{1.4}{*}{\textbf{PC}} & \multirow{1.4}{*}{\textbf{CB}} & \multirow{1.4}{*}{\textbf{Avg.}} & & \multirow{1.4}{*}{\textbf{LD}}& \multirow{1.4}{*}{\textbf{LB}} & \multirow{1.4}{*}{\textbf{Avg.}} & &\\
\midrule
GPT-4 & 32.31 & \textbf{10.02} & \textbf{19.19} & & 27.74 & 0.57&19.97 & & \textbf{0.20} & 3.76 &\textbf{2.21} & &  85.76 & 0.51& 58.17&&\textbf{33.04}  \\
LLaMA-3-8B & 46.87 & 23.46 &33.09&& 12.09 & \textbf{0.00} &8.63&& 15.48&12.99&14.07&&\textbf{78.39}&\textbf{0.00}& \textbf{53.02}&&33.67  \\
DeepSeek Coder-6.7B & 24.23 & 25.61 &25.04&& 15.80 & \textbf{0.00} &11.28&& 17.52&17.21&17.35&&99.06&0.17& 67.05&&38.28  \\
GPT-3.5 & \textbf{20.19} & 22.05 &21.28 && 30.54 & \textbf{0.00}& 21.80&& 18.53&6.42&11.68&&99.88&\textbf{0.00}&67.55 &&38.98  \\
Claude-3-haiku & 38.68 & 29.25 & 33.13&& \textbf{9.07} & 0.75 & \textbf{6.69} &&  37.07 & 20.81 & 27.88&& 100.00 &0.17&67.69  && 41.00 \\
ChatGLM-3-6B & 36.13 & 44.91 & 41.30&& 50.87 & \textbf{0.00}& 36.32&& 24.85&2.35&12.12&&88.99&\textbf{0.00}& 60.19 &&44.23 \\
Ernie-3.5& 48.14 & 36.90 & 41.52&& 30.31 & 0.38& 21.75&& 18.13&11.89&14.60&&98.98&\textbf{0.00}&  66.94 &&44.31\\
Qwen-turbo & 49.63 & 48.33 & 48.86&& 29.48 & 2.08& 21.64&& 7.94&2.82&5.04&&98.08&0.17& 66.39&&44.74\\
MagicCoder-7B  & 50.27 & 26.58 & 36.32&& 17.69 &\textbf{0.00}& 12.63&& 21.18&28.33&25.22&&100.00& 16.25& 72.90 &&44.84  \\
Code LLaMA-7B & 65.04 & 42.17 & 51.57&& 31.07 & \textbf{0.00}& 22.18&& 18.53&6.26&11.59&&94.76&9.41&67.14 &&46.68 \\
StarCoder-16B & 48.14 & 38.83 & 42.66&& 60.70 & 9.25& 45.98&& 28.92&11.11&18.85&&95.09&0.77&  64.56&&49.23  \\
LLaMA-2-7B & 51.22 & 32.29 & 40.08&& 78.46 & 71.13&76.36& & 14.87&\textbf{0.00}&6.46&&81.05&0.34& 54.93 &&49.41 \\
Gemini-1.0 & 34.11 & 53.53 & 45.54&& 45.88 & \textbf{0.00}& 32.76&& 24.44&16.12&19.73&&98.65&10.35& 70.07 &&49.57  \\
Mistral-7B & 45.48 & 36.53 &40.21&& 59.18 & 15.85& 46.79&& 27.49&10.80&18.05&&99.35&0.17 & 67.25   &&49.76 \\
WizardCoder-7B & 26.57 & 31.40 & 29.41&& 31.29 & \textbf{0.00}& 22.34&& 33.20&9.39&19.73&&93.90&72.37& 86.93&&50.10  \\
CodeGeeX-2-6B & 47.61 & 27.99 & 36.06&& 45.05 & \textbf{0.00}& 32.16&& 36.66&23.47&29.20&&89.60&99.66&92.86 &&57.47  \\
Gemma-7B & 55.26 & 41.05 & 46.90&& 51.85 & \textbf{0.00}& 37.02&& 14.46&14.55&14.51&&97.18&100.00&  98.09&&61.53 \\
\bottomrule
\end{tabular}
\end{adjustbox}
\caption{Evaluation results of 17 models on CodeHalu. \textbf{DC} denotes Data Compliance hallucination. \textbf{SA} denotes Structure Access hallucination. \textbf{ID} denotes identity hallucination. \textbf{ES} denotes External Source hallucination. \textbf{PC} denotes Physical Constraint hallucination. \textbf{CB} denotes computational Boundary hallucination. \textbf{LD} denotes Logic Deviation. \textbf{LB} denotes Logic Breakdown.}
\label{table:bench}
\end{table*}

In the validation phase, we use the CodeHalu algorithm to identify multiple types of hallucinations \(\mathsf{HaluTypes}\) $\xi$, represented as [($\xi_1$, $P_1$), \ldots, ($\xi_i$, $P_i$)]. In the identification phase, we annotate the ${k^2}$ most common hallucinations and their frequencies in each sample $\alpha_i$, represented as [($\xi_1$, $P_1$), \ldots, ($\xi_{k^2}$, $P_{k^2}$)]. In the construction phase, we sort all samples in descending order based on the frequency $P_i$ of each hallucination type $\xi_i$. If the hallucination frequency in a sample $\alpha_i$ exceeds the threshold \(k\), we include this sample in the corresponding hallucination type set in the CodeHaluEval benchmark. When selecting the threshold \(k\), we consider both the minimum number of samples required to detect code hallucination effects in the CodeHaluEval benchmark and the inference costs associated with evaluating various LLMs. 
Through this method, we establish the CodeHaluEval benchmark, with detailed statistics shown in Table \ref{table:codehalueval}.

\section{Experiments}
\label{sec:exp}

\textbf{Models.} To comprehensively analyze the different hallucinations of various competitive LLMs in CodeHaluEval, we evaluate \textbf{12 general LLMs}, including GPT-4~\citep{openai2023gpt4}, GPT-3.5~\citep{openai2023gpt4}, Gemini-Pro-1.0~\citep{team2023gemini}, Claude-3-haiku~\citep{anthropic}, LLaMA-2 \& 3~\citep{touvron2023llama}, Vicuna~\citep{chiang2023vicuna}, Qwen-turbo~\citep{bai2023qwen}, ChatGLM3-6B~\citep{du2021glm}, Ernie-3.5~\citep{ERNIE}, Mistral-7B~\citep{jiang2023mistral}, Gemma~\citep{team2024gemma}. We also evaluate \textbf{5 coding LLMs}, including Code LLaMA~\citep{roziere2023code}, DeepSeek Coder~\citep{guo2024deepseek}, CodeGeeX-2~\citep{zheng2023codegeex}, StarCoder-2~\citep{li2023starcoder}, MagicCoder-7B~\citep{wei2023magicoder}, WizardCoder-7B~\citep{luo2023wizardcoder}. 
The experimental evaluation is conducted using API calls or 8 NVIDIA A6000 GPUs.

\vspace{0.5em}

\noindent \textbf{Metrics.} Given the limited exploration of code hallucinations, no dedicated metrics currently exist for evaluating them in LLMs. To address this gap, we propose an evaluation metric called \textbf{Hallucination Rate (HR)}. Specifically, HR is defined as the percentage of hallucination samples detected in the test set among all samples, with the formula: \(\operatorname{HR} = \frac{1}{N} \sum_{i=1}^{N} S(i, K)\), where \(S(i, K)\) is an indicator function.  If the $i^{th}$ sample satisfies the hallucination condition, then $S (i, K)$ = 1; otherwise, $S (i, K)$ = 0. Ideally, a lower HR indicates a lower likelihood of hallucinations during code generation by the LLM, thus demonstrating greater robustness and reliability. To our knowledge, HR is the first metric that accurately reflects the hallucination phenomenon in LLMs during code generation tasks through actual execution tests.

\subsection{Result \& Analysis}
\label{sec:result_analysis}

\begin{figure}[t]
    \centering
    \includegraphics[width=1\linewidth]{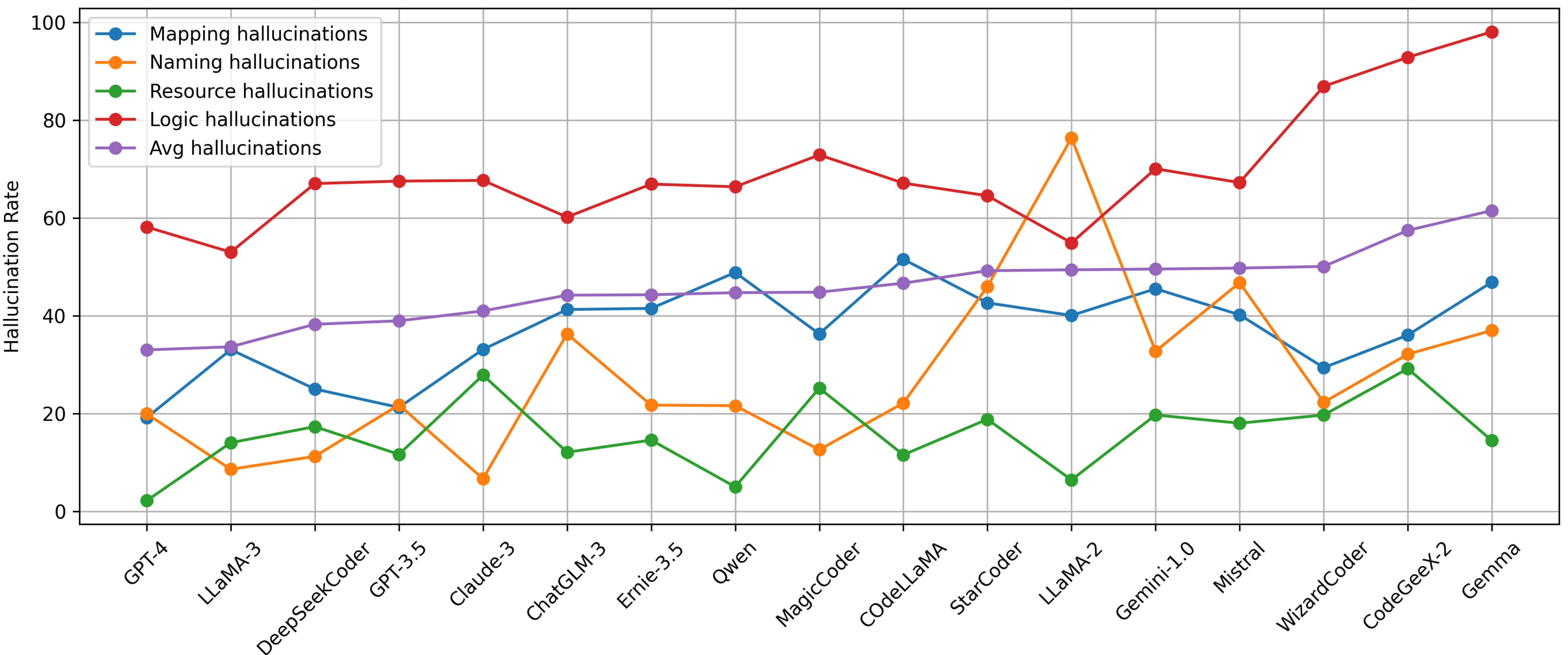}
    \caption{The performance of 17 LLMs on different types of hallucinations and the overall hallucination rate.}
    \label{fig:hallucination_result}
\end{figure}

The experimental results are presented in Table \ref{table:bench} \& Figure \ref{fig:hallucination_result}.

\textit{Mapping hallucination}: GPT-4 and GPT-3.5 consistently identify and follow rules related to data types, values, and structures, demonstrating strong context sensitivity.

\textit{Naming hallucination}: Claude-3 reliably remembers and references entity names from the context and external knowledge bases. In contrast, LLaMA-2 exhibits significant memory bias when processing external knowledge and occasionally fabricates information.

\textit{Resource hallucination}: GPT-4, Qwen, and LLaMA-2 effectively account for actual resource constraints when generating code, showing an understanding of computational boundaries and limitations, which leads them to produce code with lower complexity. 

\textit{Logical hallucination}: Although all models face challenges in maintaining logical coherence, LLaMA-3 and GPT-4 perform relatively well in reducing repetition. Most models rarely generate code with stuttering or infinite loops, but such issues are more common in Gemma, CodeGeeX-2, and WizardCoder, indicating a tendency to lose semantic and logical consistency during code generation.

\textit{Overall}, GPT-4 and LLaMA-3 perform well across all hallucination categories, displaying stability and robustness in various scenarios. Logical hallucinations remain the most prevalent issue across all models, while naming and resource hallucinations are relatively less common. The performance of different models varies significantly across hallucination types, likely due to differences in their training data, methods, and architectures. The average hallucination rate ranges from approximately 20\% to 60\%.

We view mitigating code hallucination as future work. Based on a detailed analysis of experimental results and generated cases, we provide insights into strategies for mitigating code hallucinations in LLMs. In terms of training data, improving the quality and increasing the diversity of data sources enhances the model's generalization ability. In terms of training methods, employing alignment strategies based on compilation and execution verification, as well as setting multiple objectives during training, enables the model to better understand the data flow and control flow of code. In terms of model architecture, introducing a static code verification module provides real-time feedback on verification results, thereby enhancing the model's robustness. Additionally, incorporating a code graph module allows the model to construct and utilize graph structure information when generating code, deepening its understanding of patterns and logical relationships in the generated code.

\section{Conclusion}
\label{sec:conclusion}

We introduce the concept of code hallucination and propose an execution-based verification method to classify code hallucinations. We develop the dynamic detection algorithm, CodeHalu, and categorize code hallucinations into four main types, providing a comprehensive understanding of the various challenges faced by LLMs in code generation. Additionally, we establish the CodeHaluEval benchmark and evaluate 17 widely-used LLMs, revealing significant differences in their hallucination patterns during code generation, and providing detailed insights for further improving the code generation capabilities of LLMs. Overall, we lay the theoretical foundation for understanding the hallucination phenomenon of LLMs in code generation, and provide a complete set of tools for detecting and evaluating code hallucinations.

\section{Limitations}
\label{sec:limitations}

Python is our focus for exploring code hallucination, as it is the most widely used programming language according to the TIOBE Index. Furthermore, many existing studies, such as HumanEval and MBPP benchmarks, concentrate on Python. Thus, we do not extend our investigation to other languages.

CodeHalu focuses on ensuring the correctness of generated code to meet the needs of developers and users. In contrast, identifying and preventing security risks is a higher-level concern, effectively addressed through sandbox environments. 

We focus on code hallucination specifically within the code generation task, 
but our hallucination detection and evaluation methods can be easily adapted to other tasks.

\bibliography{aaai25}

\begin{thebibliography}{43}
\providecommand{\natexlab}[1]{#1}

\bibitem[{Anthropic(2024)}]{anthropic}
Anthropic. 2024.
\newblock {T}he {C}laude 3 {M}odel {F}amily: {O}pus, {S}onnet, {H}aiku.
\newblock \url{https://www-cdn.anthropic.com/de8ba9b01c9ab7cbabf5c33b80b7bbc618857627/Model_Card_Claude_3.pdf}.

\bibitem[{Austin et~al.(2021)Austin, Odena, Nye, Bosma, Michalewski, Dohan, Jiang, Cai, Terry, Le, and Sutton}]{austin2021program}
Austin, J.; Odena, A.; Nye, M.~I.; Bosma, M.; Michalewski, H.; Dohan, D.; Jiang, E.; Cai, C.~J.; Terry, M.; Le, Q.~V.; and Sutton, C. 2021.
\newblock Program Synthesis with Large Language Models.
\newblock \emph{CoRR}, abs/2108.07732.

\bibitem[{Bai et~al.(2023)Bai, Bai, Chu, Cui, Dang, Deng, Fan, Ge, Han, Huang et~al.}]{bai2023qwen}
Bai, J.; Bai, S.; Chu, Y.; Cui, Z.; Dang, K.; Deng, X.; Fan, Y.; Ge, W.; Han, Y.; Huang, F.; et~al. 2023.
\newblock Qwen technical report.
\newblock \emph{arXiv preprint arXiv:2309.16609}.

\bibitem[{Baidu(2023)}]{ERNIE}
Baidu. 2023.
\newblock Introducing ERNIE 3.5: Baidu’s Knowledge-Enhanced Foundation Model Takes a Giant Leap Forward.

\bibitem[{Chen et~al.(2021)Chen, Tworek, Jun, Yuan, de~Oliveira~Pinto, Kaplan, Edwards, Burda, Joseph, Brockman, Ray, Puri, Krueger, Petrov, Khlaaf, Sastry, Mishkin, Chan, Gray, Ryder, Pavlov, Power, Kaiser, Bavarian, Winter, Tillet, Such, Cummings, Plappert, Chantzis, Barnes, Herbert{-}Voss, Guss, Nichol, Paino, Tezak, Tang, Babuschkin, Balaji, Jain, Saunders, Hesse, Carr, Leike, Achiam, Misra, Morikawa, Radford, Knight, Brundage, Murati, Mayer, Welinder, McGrew, Amodei, McCandlish, Sutskever, and Zaremba}]{chen2021evaluating}
Chen, M.; Tworek, J.; Jun, H.; Yuan, Q.; de~Oliveira~Pinto, H.~P.; Kaplan, J.; Edwards, H.; Burda, Y.; Joseph, N.; Brockman, G.; Ray, A.; Puri, R.; Krueger, G.; Petrov, M.; Khlaaf, H.; Sastry, G.; Mishkin, P.; Chan, B.; Gray, S.; Ryder, N.; Pavlov, M.; Power, A.; Kaiser, L.; Bavarian, M.; Winter, C.; Tillet, P.; Such, F.~P.; Cummings, D.; Plappert, M.; Chantzis, F.; Barnes, E.; Herbert{-}Voss, A.; Guss, W.~H.; Nichol, A.; Paino, A.; Tezak, N.; Tang, J.; Babuschkin, I.; Balaji, S.; Jain, S.; Saunders, W.; Hesse, C.; Carr, A.~N.; Leike, J.; Achiam, J.; Misra, V.; Morikawa, E.; Radford, A.; Knight, M.; Brundage, M.; Murati, M.; Mayer, K.; Welinder, P.; McGrew, B.; Amodei, D.; McCandlish, S.; Sutskever, I.; and Zaremba, W. 2021.
\newblock Evaluating Large Language Models Trained on Code.
\newblock \emph{CoRR}, abs/2107.03374.

\bibitem[{Cheng et~al.(2023)Cheng, Sun, Zhang, Wang, Liu, Zhang, He, Huang, Yin, Chen et~al.}]{cheng2023evaluating}
Cheng, Q.; Sun, T.; Zhang, W.; Wang, S.; Liu, X.; Zhang, M.; He, J.; Huang, M.; Yin, Z.; Chen, K.; et~al. 2023.
\newblock Evaluating hallucinations in chinese large language models.
\newblock \emph{arXiv preprint arXiv:2310.03368}.

\bibitem[{Chiang et~al.(2023)Chiang, Li, Lin, Sheng, Wu, Zhang, Zheng, Zhuang, Zhuang, Gonzalez et~al.}]{chiang2023vicuna}
Chiang, W.-L.; Li, Z.; Lin, Z.; Sheng, Y.; Wu, Z.; Zhang, H.; Zheng, L.; Zhuang, S.; Zhuang, Y.; Gonzalez, J.~E.; et~al. 2023.
\newblock Vicuna: An open-source chatbot impressing gpt-4 with 90\%* chatgpt quality.
\newblock \emph{See https://vicuna. lmsys. org (accessed 14 April 2023)}.

\bibitem[{Chu et~al.(2024)Chu, Zhang, Sun, Xue, Wang, Qin, and Ren}]{chu2024sora}
Chu, Z.; Zhang, L.; Sun, Y.; Xue, S.; Wang, Z.; Qin, Z.; and Ren, K. 2024.
\newblock Sora Detector: {A} Unified Hallucination Detection for Large Text-to-Video Models.
\newblock \emph{CoRR}, abs/2405.04180.

\bibitem[{Du et~al.(2021)Du, Qian, Liu, Ding, Qiu, Yang, and Tang}]{du2021glm}
Du, Z.; Qian, Y.; Liu, X.; Ding, M.; Qiu, J.; Yang, Z.; and Tang, J. 2021.
\newblock Glm: General language model pretraining with autoregressive blank infilling.
\newblock \emph{arXiv preprint arXiv:2103.10360}.

\bibitem[{Elaraby et~al.(2023)Elaraby, Lu, Dunn, Zhang, Wang, and Liu}]{elaraby2023halo}
Elaraby, M.; Lu, M.; Dunn, J.; Zhang, X.; Wang, Y.; and Liu, S. 2023.
\newblock Halo: Estimation and Reduction of Hallucinations in Open-Source Weak Large Language Models.
\newblock \emph{CoRR}, abs/2308.11764.

\bibitem[{Gemini(2023)}]{team2023gemini}
Gemini. 2023.
\newblock Gemini: a family of highly capable multimodal models.
\newblock \emph{arXiv preprint arXiv:2312.11805}.

\bibitem[{Guo et~al.(2024)Guo, Zhu, Yang, Xie, Dong, Zhang, Chen, Bi, Wu, Li et~al.}]{guo2024deepseek}
Guo, D.; Zhu, Q.; Yang, D.; Xie, Z.; Dong, K.; Zhang, W.; Chen, G.; Bi, X.; Wu, Y.; Li, Y.; et~al. 2024.
\newblock DeepSeek-Coder: When the Large Language Model Meets Programming--The Rise of Code Intelligence.
\newblock \emph{arXiv preprint arXiv:2401.14196}.

\bibitem[{Hendrycks et~al.(2021)Hendrycks, Basart, Kadavath, Mazeika, Arora, Guo, Burns, Puranik, He, Song et~al.}]{hendrycks2021measuring}
Hendrycks, D.; Basart, S.; Kadavath, S.; Mazeika, M.; Arora, A.; Guo, E.; Burns, C.; Puranik, S.; He, H.; Song, D.; et~al. 2021.
\newblock Measuring coding challenge competence with apps.
\newblock \emph{arXiv preprint arXiv:2105.09938}.

\bibitem[{Huang et~al.(2023)Huang, Meng, Zhang, Liu, Wang, Li, and Zhang}]{10298532}
Huang, K.; Meng, X.; Zhang, J.; Liu, Y.; Wang, W.; Li, S.; and Zhang, Y. 2023.
\newblock An Empirical Study on Fine-Tuning Large Language Models of Code for Automated Program Repair.
\newblock In \emph{2023 38th IEEE/ACM International Conference on Automated Software Engineering (ASE)}, 1162--1174.

\bibitem[{Huang et~al.(2024)Huang, Liu, Guo, and Gong}]{huang2024visual}
Huang, W.; Liu, H.; Guo, M.; and Gong, N.~Z. 2024.
\newblock Visual Hallucinations of Multi-modal Large Language Models.
\newblock \emph{CoRR}, abs/2402.14683.

\bibitem[{Jesse et~al.(2023)Jesse, Ahmed, Devanbu, and Morgan}]{10174227}
Jesse, K.; Ahmed, T.; Devanbu, P.~T.; and Morgan, E. 2023.
\newblock Large Language Models and Simple, Stupid Bugs.
\newblock In \emph{2023 IEEE/ACM 20th International Conference on Mining Software Repositories (MSR)}, 563--575. Los Alamitos, CA, USA: IEEE Computer Society.

\bibitem[{Ji et~al.(2023)Ji, Lee, Frieske, Yu, Su, Xu, Ishii, Bang, Madotto, and Fung}]{ji2023survey}
Ji, Z.; Lee, N.; Frieske, R.; Yu, T.; Su, D.; Xu, Y.; Ishii, E.; Bang, Y.~J.; Madotto, A.; and Fung, P. 2023.
\newblock Survey of hallucination in natural language generation.
\newblock \emph{ACM Computing Surveys}, 55(12): 1--38.

\bibitem[{Jiang et~al.(2023)Jiang, Sablayrolles, Mensch, Bamford, Chaplot, Casas, Bressand, Lengyel, Lample, Saulnier et~al.}]{jiang2023mistral}
Jiang, A.~Q.; Sablayrolles, A.; Mensch, A.; Bamford, C.; Chaplot, D.~S.; Casas, D. d.~l.; Bressand, F.; Lengyel, G.; Lample, G.; Saulnier, L.; et~al. 2023.
\newblock Mistral 7B.
\newblock \emph{arXiv preprint arXiv:2310.06825}.

\bibitem[{Jimenez et~al.(2023)Jimenez, Yang, Wettig, Yao, Pei, Press, and Narasimhan}]{jimenez2023swe}
Jimenez, C.~E.; Yang, J.; Wettig, A.; Yao, S.; Pei, K.; Press, O.; and Narasimhan, K. 2023.
\newblock Swe-bench: Can language models resolve real-world github issues?
\newblock \emph{arXiv preprint arXiv:2310.06770}.

\bibitem[{Li et~al.(2024)Li, Tian, Hu, Luo, and Ma}]{li2024mmcode}
Li, K.; Tian, Y.; Hu, Q.; Luo, Z.; and Ma, J. 2024.
\newblock MMCode: Evaluating Multi-Modal Code Large Language Models with Visually Rich Programming Problems.
\newblock arXiv:2404.09486.

\bibitem[{Li et~al.(2023{\natexlab{a}})Li, Allal, Zi, Muennighoff, Kocetkov, Mou, Marone, Akiki, Li, Chim et~al.}]{li2023starcoder}
Li, R.; Allal, L.~B.; Zi, Y.; Muennighoff, N.; Kocetkov, D.; Mou, C.; Marone, M.; Akiki, C.; Li, J.; Chim, J.; et~al. 2023{\natexlab{a}}.
\newblock Starcoder: may the source be with you!
\newblock \emph{arXiv preprint arXiv:2305.06161}.

\bibitem[{Li et~al.(2023{\natexlab{b}})Li, Fu, Zhang, Huang, Sun, Lyu, Liu, Jin, and Li}]{li2023taco}
Li, R.; Fu, J.; Zhang, B.-W.; Huang, T.; Sun, Z.; Lyu, C.; Liu, G.; Jin, Z.; and Li, G. 2023{\natexlab{b}}.
\newblock Taco: Topics in algorithmic code generation dataset.
\newblock \emph{arXiv preprint arXiv:2312.14852}.

\bibitem[{Lin, Hilton, and Evans(2021)}]{lin2021truthfulqa}
Lin, S.; Hilton, J.; and Evans, O. 2021.
\newblock Truthfulqa: Measuring how models mimic human falsehoods.
\newblock \emph{arXiv preprint arXiv:2109.07958}.

\bibitem[{Liu et~al.(2023)Liu, Lin, Li, Wang, Yacoob, and Wang}]{liu2023aligning}
Liu, F.; Lin, K.; Li, L.; Wang, J.; Yacoob, Y.; and Wang, L. 2023.
\newblock Aligning large multi-modal model with robust instruction tuning.
\newblock \emph{arXiv preprint arXiv:2306.14565}.

\bibitem[{Liu et~al.(2024)Liu, Xue, Chen, Chen, Zhao, Wang, Hou, Li, and Peng}]{liu2024survey}
Liu, H.; Xue, W.; Chen, Y.; Chen, D.; Zhao, X.; Wang, K.; Hou, L.; Li, R.; and Peng, W. 2024.
\newblock A survey on hallucination in large vision-language models.
\newblock \emph{arXiv preprint arXiv:2402.00253}.

\bibitem[{Luo et~al.(2023)Luo, Xu, Zhao, Sun, Geng, Hu, Tao, Ma, Lin, and Jiang}]{luo2023wizardcoder}
Luo, Z.; Xu, C.; Zhao, P.; Sun, Q.; Geng, X.; Hu, W.; Tao, C.; Ma, J.; Lin, Q.; and Jiang, D. 2023.
\newblock Wizardcoder: Empowering code large language models with evol-instruct.
\newblock \emph{arXiv preprint arXiv:2306.08568}.

\bibitem[{OpenAI(2023)}]{openai2023gpt4}
OpenAI. 2023.
\newblock {GPT-4} Technical Report.

\bibitem[{Pan et~al.(2023)Pan, Ibrahimzada, Krishna, Sankar, Wassi, Merler, Sobolev, Pavuluri, Sinha, and Jabbarvand}]{pan2023understanding}
Pan, R.; Ibrahimzada, A.~R.; Krishna, R.; Sankar, D.; Wassi, L.~P.; Merler, M.; Sobolev, B.; Pavuluri, R.; Sinha, S.; and Jabbarvand, R. 2023.
\newblock Understanding the effectiveness of large language models in code translation.
\newblock \emph{arXiv preprint arXiv:2308.03109}.

\bibitem[{Peng et~al.(2023)Peng, Galley, He, Cheng, Xie, Hu, Huang, Liden, Yu, Chen et~al.}]{peng2023check}
Peng, B.; Galley, M.; He, P.; Cheng, H.; Xie, Y.; Hu, Y.; Huang, Q.; Liden, L.; Yu, Z.; Chen, W.; et~al. 2023.
\newblock Check your facts and try again: Improving large language models with external knowledge and automated feedback.
\newblock \emph{arXiv preprint arXiv:2302.12813}.

\bibitem[{Roziere et~al.(2023)Roziere, Gehring, Gloeckle, Sootla, Gat, Tan, Adi, Liu, Remez, Rapin et~al.}]{roziere2023code}
Roziere, B.; Gehring, J.; Gloeckle, F.; Sootla, S.; Gat, I.; Tan, X.~E.; Adi, Y.; Liu, J.; Remez, T.; Rapin, J.; et~al. 2023.
\newblock Code llama: Open foundation models for code.
\newblock \emph{arXiv preprint arXiv:2308.12950}.

\bibitem[{Team et~al.(2024)Team, Mesnard, Hardin, Dadashi, Bhupatiraju, Pathak, Sifre, Rivi{\`e}re, Kale, Love et~al.}]{team2024gemma}
Team, G.; Mesnard, T.; Hardin, C.; Dadashi, R.; Bhupatiraju, S.; Pathak, S.; Sifre, L.; Rivi{\`e}re, M.; Kale, M.~S.; Love, J.; et~al. 2024.
\newblock Gemma: Open models based on gemini research and technology.
\newblock \emph{arXiv preprint arXiv:2403.08295}.

\bibitem[{Touvron et~al.(2023)Touvron, Martin, Stone, Albert, Almahairi, Babaei, Bashlykov, Batra, Bhargava, Bhosale et~al.}]{touvron2023llama}
Touvron, H.; Martin, L.; Stone, K.; Albert, P.; Almahairi, A.; Babaei, Y.; Bashlykov, N.; Batra, S.; Bhargava, P.; Bhosale, S.; et~al. 2023.
\newblock Llama 2: Open foundation and fine-tuned chat models.
\newblock \emph{arXiv preprint arXiv:2307.09288}.

\bibitem[{Wang et~al.(2024)Wang, Zhou, Song, Huang, Chen, Ma, and Zhang}]{wang2024large}
Wang, Z.; Zhou, Z.; Song, D.; Huang, Y.; Chen, S.; Ma, L.; and Zhang, T. 2024.
\newblock Where Do Large Language Models Fail When Generating Code?
\newblock \emph{arXiv preprint arXiv:2406.08731}.

\bibitem[{Wei et~al.(2023)Wei, Wang, Liu, Ding, and Zhang}]{wei2023magicoder}
Wei, Y.; Wang, Z.; Liu, J.; Ding, Y.; and Zhang, L. 2023.
\newblock Magicoder: Source code is all you need.
\newblock \emph{arXiv preprint arXiv:2312.02120}.

\bibitem[{Xu et~al.(2020)Xu, van Genabith, Xiong, Liu, and Zhang}]{xu-etal-2020-learning}
Xu, H.; van Genabith, J.; Xiong, D.; Liu, Q.; and Zhang, J. 2020.
\newblock Learning Source Phrase Representations for Neural Machine Translation.
\newblock In Jurafsky, D.; Chai, J.; Schluter, N.; and Tetreault, J., eds., \emph{Proceedings of the 58th Annual Meeting of the Association for Computational Linguistics}, 386--396. Online: Association for Computational Linguistics.

\bibitem[{Xu et~al.(2022)Xu, Liu, Yan, Cai, Li, and Li}]{xu2022learning}
Xu, J.; Liu, X.; Yan, J.; Cai, D.; Li, H.; and Li, J. 2022.
\newblock Learning to break the loop: Analyzing and mitigating repetitions for neural text generation.
\newblock \emph{Advances in Neural Information Processing Systems}, 35: 3082--3095.

\bibitem[{Yan and Li(2022)}]{yan2022whygen}
Yan, W.; and Li, Y. 2022.
\newblock WhyGen: explaining ML-powered code generation by referring to training examples.
\newblock In \emph{Proceedings of the ACM/IEEE 44th International Conference on Software Engineering: Companion Proceedings}, 237--241.

\bibitem[{Yan et~al.(2023)Yan, Liu, Wang, Li, Chen, Wang, Lin, Zhao, Zhu, Deng et~al.}]{yan2023codescope}
Yan, W.; Liu, H.; Wang, Y.; Li, Y.; Chen, Q.; Wang, W.; Lin, T.; Zhao, W.; Zhu, L.; Deng, S.; et~al. 2023.
\newblock Codescope: An execution-based multilingual multitask multidimensional benchmark for evaluating llms on code understanding and generation.
\newblock \emph{arXiv preprint arXiv:2311.08588}.

\bibitem[{Yang, Liu, and Yin(2021)}]{e23091174}
Yang, C.; Liu, Y.; and Yin, C. 2021.
\newblock Recent Advances in Intelligent Source Code Generation: A Survey on Natural Language Based Studies.
\newblock \emph{Entropy}, 23(9).

\bibitem[{Zhai et~al.(2023)Zhai, Yang, Zhao, Xu, Shen, Zhao, Keutzer, Li, Yan, and Fan}]{zhai2023halle}
Zhai, B.; Yang, S.; Zhao, X.; Xu, C.; Shen, S.; Zhao, D.; Keutzer, K.; Li, M.; Yan, T.; and Fan, X. 2023.
\newblock Halle-switch: Rethinking and controlling object existence hallucinations in large vision language models for detailed caption.
\newblock \emph{arXiv preprint arXiv:2310.01779}.

\bibitem[{Zhang et~al.(2024)Zhang, Li, Zhang, and Jin}]{zhang2024hirope}
Zhang, K.; Li, G.; Zhang, H.; and Jin, Z. 2024.
\newblock HiRoPE: Length Extrapolation for Code Models.
\newblock \emph{arXiv preprint arXiv:2403.19115}.

\bibitem[{Zhang et~al.(2023)Zhang, Li, Cui, Cai, Liu, Fu, Huang, Zhao, Zhang, Chen et~al.}]{zhang2023siren}
Zhang, Y.; Li, Y.; Cui, L.; Cai, D.; Liu, L.; Fu, T.; Huang, X.; Zhao, E.; Zhang, Y.; Chen, Y.; et~al. 2023.
\newblock Siren's song in the AI ocean: a survey on hallucination in large language models.
\newblock \emph{arXiv preprint arXiv:2309.01219}.

\bibitem[{Zheng et~al.(2023)Zheng, Xia, Zou, Dong, Wang, Xue, Wang, Shen, Wang, Li et~al.}]{zheng2023codegeex}
Zheng, Q.; Xia, X.; Zou, X.; Dong, Y.; Wang, S.; Xue, Y.; Wang, Z.; Shen, L.; Wang, A.; Li, Y.; et~al. 2023.
\newblock Codegeex: A pre-trained model for code generation with multilingual evaluations on humaneval-x.
\newblock \emph{arXiv preprint arXiv:2303.17568}.

\end{thebibliography}

\end{document}